\documentclass[12pt]{article}\usepackage[]{graphicx}\usepackage[]{color}
\makeatletter
\def\maxwidth{ %
  \ifdim\Gin@nat@width>\linewidth
    \linewidth
  \else
    \Gin@nat@width
  \fi
}
\makeatother

\definecolor{fgcolor}{rgb}{0.345, 0.345, 0.345}

\usepackage{framed}
\makeatletter
 {\par\unskip\endMakeFramed%
 \at@end@of@kframe}
\makeatother

\definecolor{shadecolor}{rgb}{.97, .97, .97}
\definecolor{messagecolor}{rgb}{0, 0, 0}
\definecolor{warningcolor}{rgb}{1, 0, 1}
\definecolor{errorcolor}{rgb}{1, 0, 0}
\newenvironment{knitrout}{}{} 

\usepackage{alltt}
\usepackage{multirow}
\usepackage{mathtools}
\usepackage{amsfonts}
\usepackage{amsmath}
\usepackage{rotating}
\usepackage{booktabs}
\usepackage{verbatim}
\usepackage{mathrsfs}
\usepackage{hyperref}
\usepackage{pdflscape}
\usepackage{caption}
\usepackage{tablefootnote}
\usepackage{url}
\usepackage[authoryear]{natbib}
\usepackage{verbatim}
\usepackage{setspace}
\usepackage[utf8]{inputenc}
\usepackage{authblk}
\usepackage{pdfpages}
\usepackage{fancyhdr}
\newcommand{\pkg}[1]{{\fontseries{b}\selectfont #1}}
\title{A Projection Pursuit Forest Algorithm for Supervised Classification}
\author{Natalia da Silva, Dianne Cook \& Eun-Kyung Lee}
\IfFileExists{upquote.sty}{\usepackage{upquote}}{}
\begin{document}
\maketitle

\section*{Abstract}

This paper presents a new ensemble learning method for classification problems called projection pursuit random forest (PPF). PPF uses the \pkg{PPtree} algorithm introduced in \cite{lee2013pptree}. In PPF, trees are constructed by splitting on linear combinations of randomly chosen variables. Projection pursuit is used to choose a projection of the variables that best separates the classes. Utilizing linear combinations of variables to separate classes takes the correlation between variables into account which allows PPF to outperform a traditional random forest when separations between groups occurs in combinations of variables. 

The method presented here can be used in multi-class problems and is implemented into an R~\citep{RCore} package, \pkg{PPforest}, which is available on CRAN.

\newpage
\linespread{2}

\section{Introduction}
There are two main aspects of a random forest~\citep{breiman2001random}, bootstrap aggregation and \citep{breiman1996bagging, breiman1996heuristics} random predictor selection \citep{amit1997shape, ho1998random}, that are broadly applicable to build ensemble classifiers from any basic method. Bagging stabilizes the variance and random predictor selection reduces correlation between trees in the forest.

This paper presents the projection pursuit random forest (PPF), a new ensemble learning method for classification problems, built on combinations of predictors in the tree construction.

PPF builds on the projection pursuit tree (PPtree) algorithm~\citep{lee2013pptree}, available in the R package \pkg{PPtreeViz}~\citep{PPtreeViz} which fits a single multi-class tree to the data. Projection pursuit is used to find the linear combination of variables that best separates groups, and many different rules to make the actual split are provided. 

Trees that use linear combinations of predictors in a split are known in the literature as oblique trees~\citep{kim2001,brodley,tan2005mml,truong2009fast,lee2013pptree}.
All these algorithms use different approaches for finding linear combinations of predictors upon which to make a split. Some of the methods used for selecting the linear combination include random coefficient generation, linear discriminant analysis, and linear support vector machines.
Theoretically, these could also be used as a base underlying PPF.

For each split, a random sample of predictors is selected, then an optimal linear combination for separating the classes is computed by using a projection pursuit index. The algorithm is targeted for problems where classes can be separated by linear combinations of predictors, which define separating hyperplanes that are oblique to the axes rather than orthogonal to them. Additionally PPF accommodates class imbalance by using stratified bootstrap samples and variable importance measures are computed using the coefficients of the projections. PPF can be used for multi-class problems and is implemented into an R package, called \pkg{PPforest}.  Only the LDA and PDA projection pursuit indexes are available in PPF.

In the machine learning literature numerous work has been conducted on algorithms for building forests from oblique trees~\citep{tan2006decision}, \citep{menze2011oblique}  and \citep{do2010classifying}. The performance is reported to be better than random forests, which is what we have determined with our algorithm also. A limitation of building on these approaches is the lack of readily available software.

This paper is organized as follows. Section \ref{PPT} explains the projection pursuit tree underlying PPF. Section \ref{PPFsec} describes the PPF algorithm; diagnostics, including how to compute variable importance and the implementation details. Section \ref{perfsec} evaluates the algorithm using a simulation study and performance on benchmark machine learning data in comparison with other methods. Section \ref{options} discusses the choice of parameters, and compares the diagnostics relative to random forests.  Section \ref{discpp1} discusses possible extensions and future directions.

\section{Background on the projection pursuit tree}~\label{PPT}

The projection pursuit algorithm searches for a low dimensional projection that optimizes a continuous function which measures some aspect of interest; for PPF, this is class separation. \cite{friedman1973projection} coined the term ``projection pursuit'', but the ideas existed earlier than this \citep{kruskal1969toward}.
\cite{lee2005projection} developed an index, derived from the linear discriminant analysis, for finding projections that separate classes.
Let $\mathbf{x_{gi}}$ be a $p$-dimensional data vector, $i$-th observation of the $g$-th class, $g = \{1,\ldots, G\}$, $G$ is the number of classes, $i = \{1,\ldots , n_g\}$, and $n_g$ is the number of observations in class $g$. The LDA index is defined as follows:

\begin{equation}
 I_{LDA}(A) = \left\{
  \begin{array}{l l}
    1-\frac{|A^T WA|}{|A^T(W+B)A|} &  \text{for} |A^T(W+B)A|\neq 0\\
    0 &  \text{for}  |A^T(W+B)A|= 0
  \end{array} \right.
  \end{equation}

\noindent where $B=\sum_{g=1}^G n_g(\bar{\mathbf{x}}_\mathbf{{g.}}-\bar{\mathbf{x}}_\mathbf{{..}})(\bar{\mathbf{x}}_\mathbf{{g.}}-\bar{\mathbf{x}}_\mathbf{{..}})^{T}$ is the between-group sums of squares, and $W=\sum_{g=1}^{G}\sum_{i=1}^{n_g}(\mathbf{x}_\mathbf{{ig}}-\bar{\mathbf{x}}_\mathbf{{g.}})(\mathbf{x}_\mathbf{{ig}}-\bar{\mathbf{x}}_\mathbf{{.g}})^T$ is the within-group sums of squares. If the LDA index value is high, there is a large difference between classes.

A second index, PDA, was developed to address large $p$, small $n$ data~\citep{lee2010projection}. The main idea used in construction of the index is that when $n\leq p$ or the variables are highly correlated,  the maximum likelihood variance-covariance matrix estimator will be close to being singular, and this will affect the inverse calculation. The PDA index adjusts the variance-covariance matrix calculation, and is defined as follows:

\begin{equation}
I_{PDA}(A,\lambda)=1-\frac{|A^T W_{PDA}A|}{|A^T (W_{PDA}+B) A|}
\end{equation}

\noindent where A is an orthonormal projection onto a $k$-dimensional space and $\lambda \in [0,1)$ is a pre-determined parameter. $B$ is the between-class sums of squares  and $W_{PDA}=\mbox{diag}(W)+(1-\lambda)\mbox{offdiag}(W)$.

\begin{figure}[!t]
\begin{knitrout}
\definecolor{shadecolor}{rgb}{0.969, 0.969, 0.969}\color{fgcolor}
\includegraphics[width=\maxwidth]{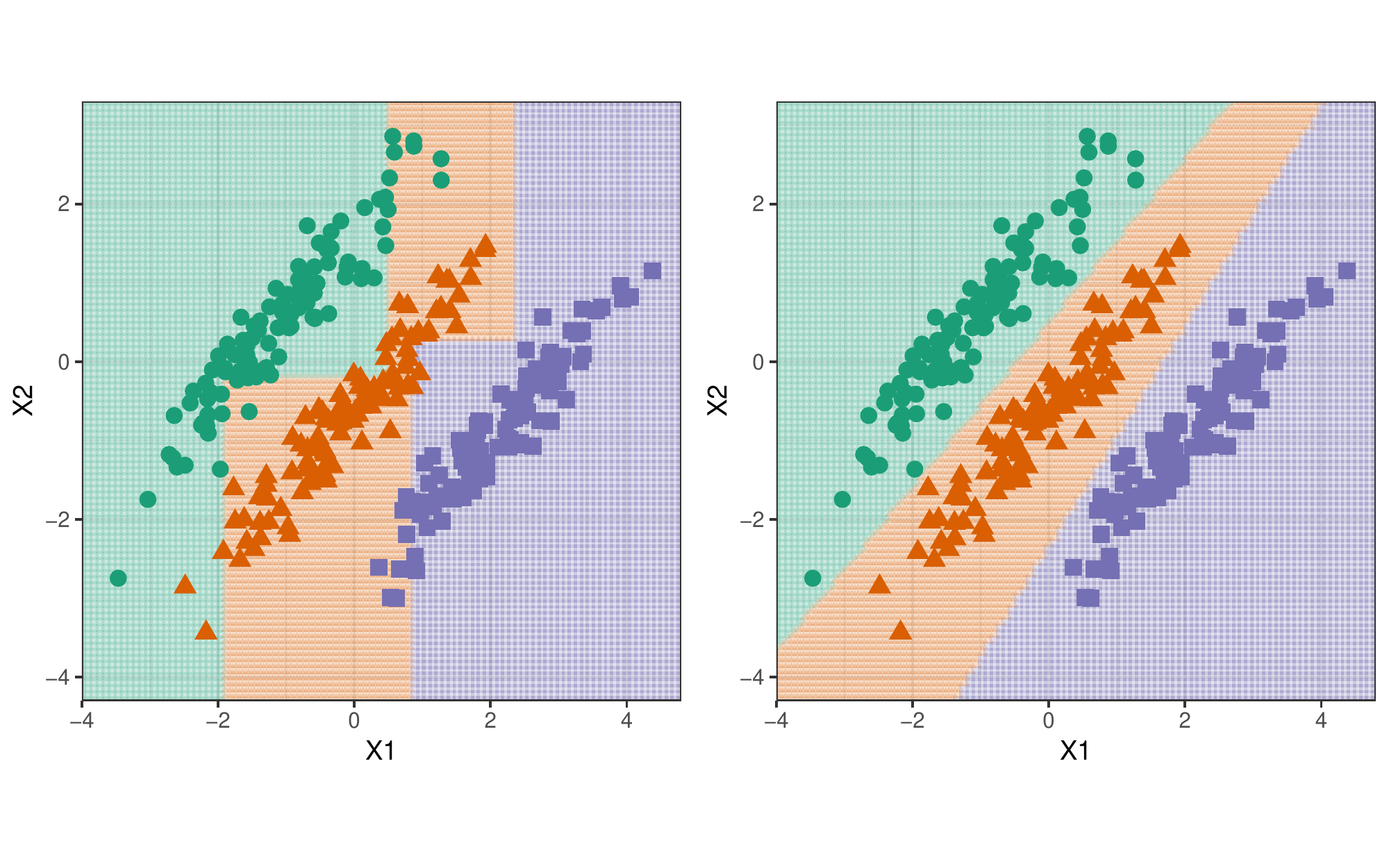} 

\end{knitrout}
 \vspace*{-0.3cm}
 \caption{Comparison of decision boundaries for the \pkg{rpart} (left) and PPtree (right) algorithms on 2D simulated data. The partitions generated by PPtree algorithm are oblique to the axis, incorporating the association between the two variables.\label{bounds}}
\end{figure}

The PPtree algorithm uses a multi-step approach to fit a multi-class model by finding linear combinations to split on. Figure \ref{bounds} compares the boundaries that would result from a classification tree fitted using the rpart algorithm~\citep{therneau2010rpart} and the PPtree algorithm.

\begin{figure}[!hbt]
\centering
\includegraphics[width=1\linewidth]{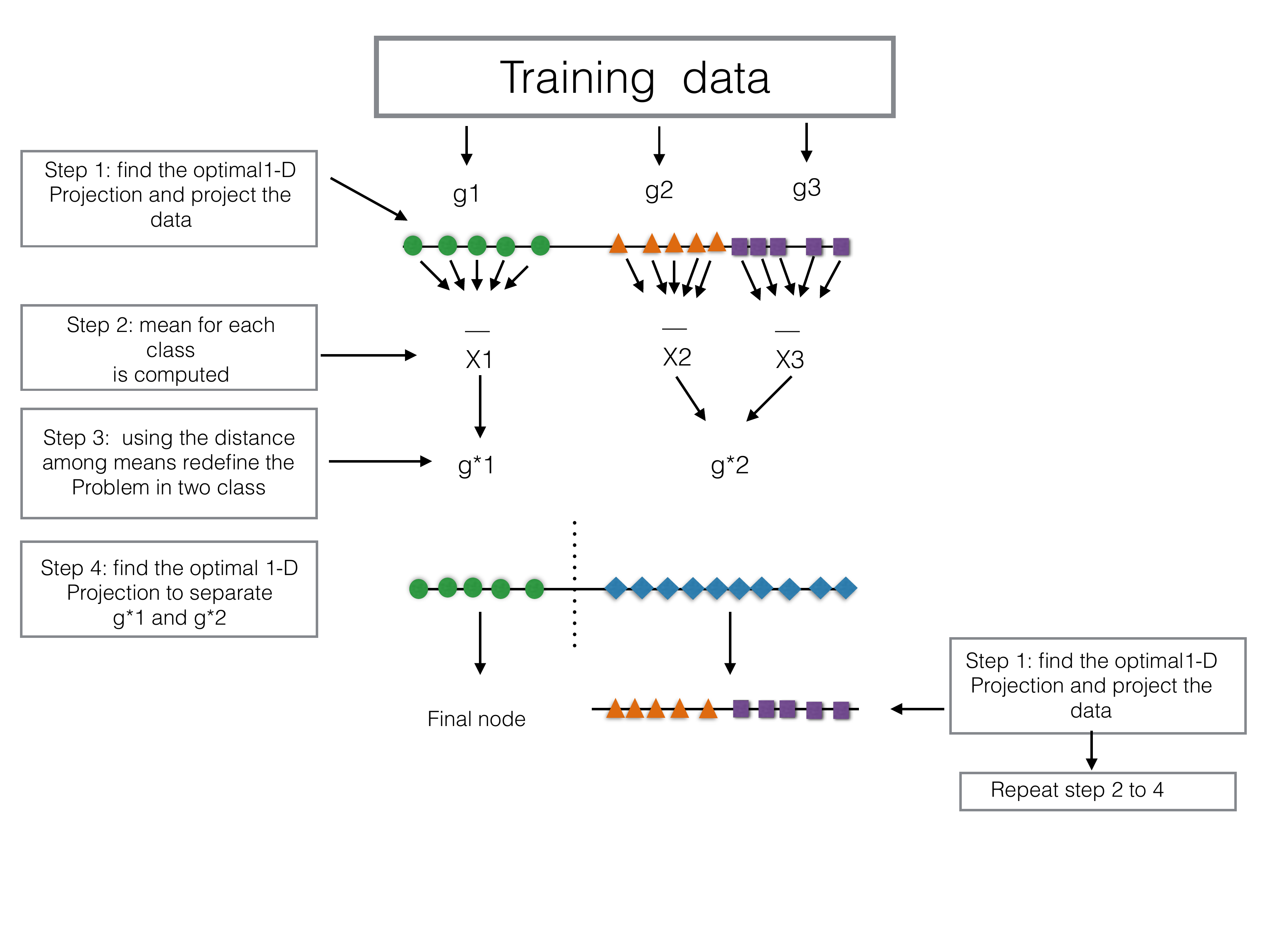}

\caption{Illustration of the PPtree algorithm for $g=3$ classes. It is a dual pass algorithm for multiclass problems, for each split. It first finds the best separation and combines classes into two super-groups. It then searches again for the best separation between these two super-groups and splits on this. It proceeds sequentially on the subsets in the nodes, but only $g-1$ splits are allowed. \label{diagpp1}}
\end{figure}

Figure \ref{diagpp1} illustrates the PPtree algorithm for three classes, and the algorithm steps are detailed below. Let  $d_n =\{(\mathbf{x_i},y_i)\}_{i=1}^n$ be the data set where $\mathbf{x_i}$ is a  p-dimensional vector of explanatory variables and  $y_i\in \mathscr{G}$ ($\mathscr{G} =\{1,2,\ldots G\}$) represents class information with $i=1,\ldots n$.

\begin{enumerate}
\item Optimize a projection pursuit index to find an optimal one-dimensional projection, $\alpha^*$, for separating all classes in the current data yielding projected data $z = \alpha^*x$.
\item On the projected data, $z$, redefine the problem into a two class problem by comparing means, and assign a new label, either $g_1^*$ or $g_2^*$ to each observation, generating a new class variable $y_i^*$.  The new groups $g_1^*$ and $g_2^*$ can contain more than one original class.
\item Find an optimal one-dimensional projection $\alpha^{**}$, using $\{(\mathbf{x_i},y_i^*)\}_{i=1}^n$ to separate the two class problem $g_1^*$ and $g_2^*$. The best separation of $g_1^*$ and $g_2^*$ is determined in this step providing the decision rule for the node,

\begin{quote}
if $\alpha^{**T}M_1< c$ then assign $g_1^*$ to the left node else assign $g_2^*$ to the right node,
\end{quote}
\noindent where $M_1$ is the mean of $g_1^*$.
\item For each group, all the previous steps are repeated until $g_1^*$ and $g_2^*$ have only one class from the original classes. The depth of PPtree is at most the number of classes.
\end{enumerate}

\newpage

\section{Projection pursuit random forest}\label{PPFsec}

This section provides the definition of PPF for classification and the algorithm. Diagnostics for the classifier are also defined.

\subsection{Definition}

Let the random vector of predictor variables $\mathbf{X}\in R^p$  and  the output random variable $Y \in \mathscr{G}$, where $\mathscr{G}$ is a finite set such that $\mathscr{G}=\{1,2, \ldots, G\}$. The training sample is defined as $D_n=\{(\mathbf{X_1}, Y_1), \ldots (\mathbf{X}_n, Y_n)\}$ of i.i.d $\Re^p \times \mathscr{G}$ random variables $(p\geq 2)$.
The objective is to build a classifier which predicts $y$ from $\mathbf{x}$ using $D_n$ given an ensemble of classifiers $h$.

 A projection pursuit classification random forest can be defined as a collection of randomized classification trees $\{h_n(\mathbf{x}, \Theta_m, D_n), m\geq 1\}$  where $\{\Theta_m\}$ are i.i.d.~random vectors. $\Theta_m$ includes the two sources of randomness in the tree (random variable selection and random bootstrap sample), then $\Theta_m$ has information about which variables were selected in each partition and which cases were selected in the bootstrap sample.

For each tree, $h_n$, a unique vote is collected based on the most popular class for the selected predictor variables.  Equation \ref{rfesti} defines the PPF estimator based on combining the trees.

\begin{eqnarray}\label{rfesti}
f_n(\mathbf{X}, D_n )&=& \operatorname*{arg\,max}_{g\in \mathscr{G}} \{E_{\Theta}(I[h_n(\mathbf{X}, \Theta, D_n)=g])\}\\ \nonumber
&=& \operatorname*{arg\,max}_{g\in \mathscr{G}} P_{\Theta}(h_n(\mathbf{X}, \Theta, D_n)=g)
\end{eqnarray}

\noindent $E_{\Theta}$ is the expectation wrt $\Theta$, conditionally on $\mathbf{X}$ and $D_n$.
In practice, the PPF estimator is evaluated by generating $B$ random trees and take the average of the individual outcomes. This procedure is justified in a similar way to the original random forest defined by \cite{breiman2001random}, and is based on the Law of the Large Numbers~\citep{athreya2006measure}.

Equation \ref{predfor} describes the prediction of a new observation $\mathbf{x_0}$.

\begin{equation}
\hat f_n(\mathbf{x_0})= \operatorname*{arg\,max}_{g\in \mathscr{G}} \sum_{k=1}^B I [ h_n (\mathbf{x_0}, \Theta_{bk} )= g]
\label{predfor}
\end{equation}

\subsection{Algorithm }

\begin{enumerate}

\item Let $n=\sum_{i=1}^G n_i$ the total number of cases in the training set $d_n=\{\mathbf{x_i}, y_i\}_{i=1}^n$. $B$ stratified bootstrap samples from $d_n$ are taken. Then for each class, independently and uniformly re-sample cases from $d_{ng}$ (training data set for group $g$) with size $n_g$ to create a stratified bootstrap data set  $\{bk= b_{k1}, b_{k2}, \ldots b_{kg}\}$.

\item Use a bootstrap sample $bk$ to grow a PPtree $(h_n(\mathbf{x}, \Theta_{bk}))$ to the largest extent possible without pruning. (Note that the depth of the PPtree is at most $G-1$, where $G$ is the number of classes).

\begin{enumerate}
\item Start with all the cases in $b_k$ in the root node.
\item A simple random sample of $m$ predictor variables from the set of all the predictor variables $M$ is drawn, where $m<<M$.
\item Find the optimal one-dimensional projection $\alpha^*$ to separate all the classes in $b_k$.
\item If more than two class, then reduce the number of classes to two by comparing means, and assign new labels, $g_1^*$ and $g_2^*$ to each case (called the new response $y_i^*$ in $b_k$).

\item Find the optimal one-dimensional projection, $\alpha^{**}$, using the bootstrap data set with the relabeled response, $y^*$, to separate $g_1^*$ and $g_2^*$.
The linear combination is computed by optimizing a projection pursuit index to get a projection of the variables that best separates the classes using the $m$ random selected variables. Two index options are available LDA or PDA.
\item Compute the decision boundary $c$.
Eight different rules to define the cutoff value of each node can be used. All the rules are defined in \cite{lee2018pptreeviz}.
\item Keep $\alpha^{**}$ and $c$.
\item Separate the data into two groups using the new labels $g_1^*$ and $g_2^*$.
\item Repeat from (b) to (h) if $g_1^*$ or $g_2^*$ have more than two original classes.
\end{enumerate}
\item Repeat 2 for $k = 1,\ldots B$.
\item The output is the ensemble of PPtrees, $\{h_n^{bk}\}_{k=1}^B$.
\end{enumerate}

Split values on the projected data can be computed by one of eight methods, which use the group means, or medians, sample size and variance or IQR weighting

\begin{figure}[!ht]
\centering
\includegraphics[width=1\linewidth]{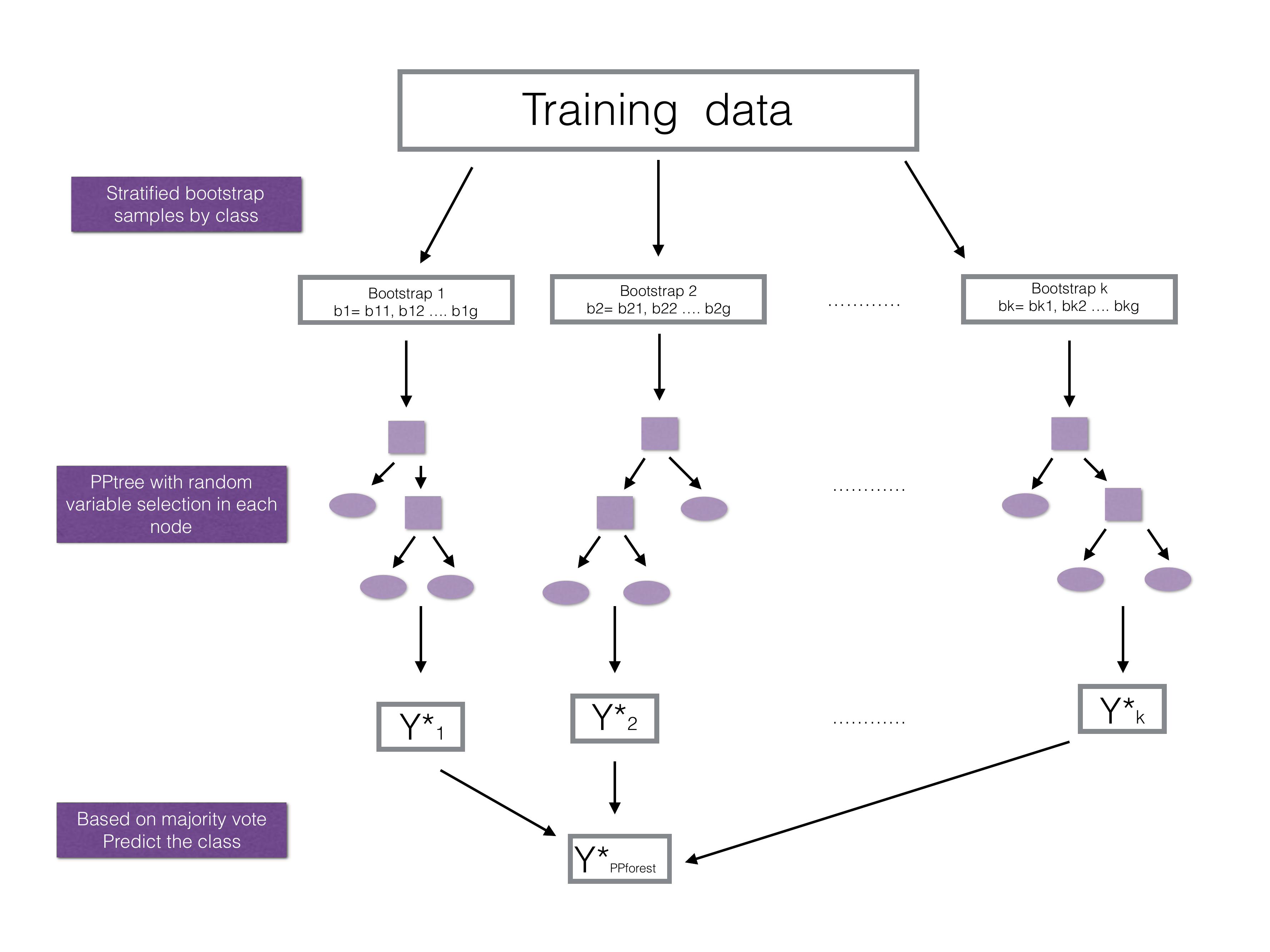}
 \vspace*{-0.5cm}
\caption{Illustration of the PPforest algorithm. It is effectively the same as a random forest algorithm except that the PPtree classifier is used on each bootstrap sample. \label{diagppf}}
\end{figure}

Figure \ref{diagppf} has a diagram illustrating the PPforest algorithm.

\subsection{Implementation}

\begin{figure}
\begin{knitrout}
\definecolor{shadecolor}{rgb}{0.969, 0.969, 0.969}\color{fgcolor}
\includegraphics[width=15cm,height=15cm]{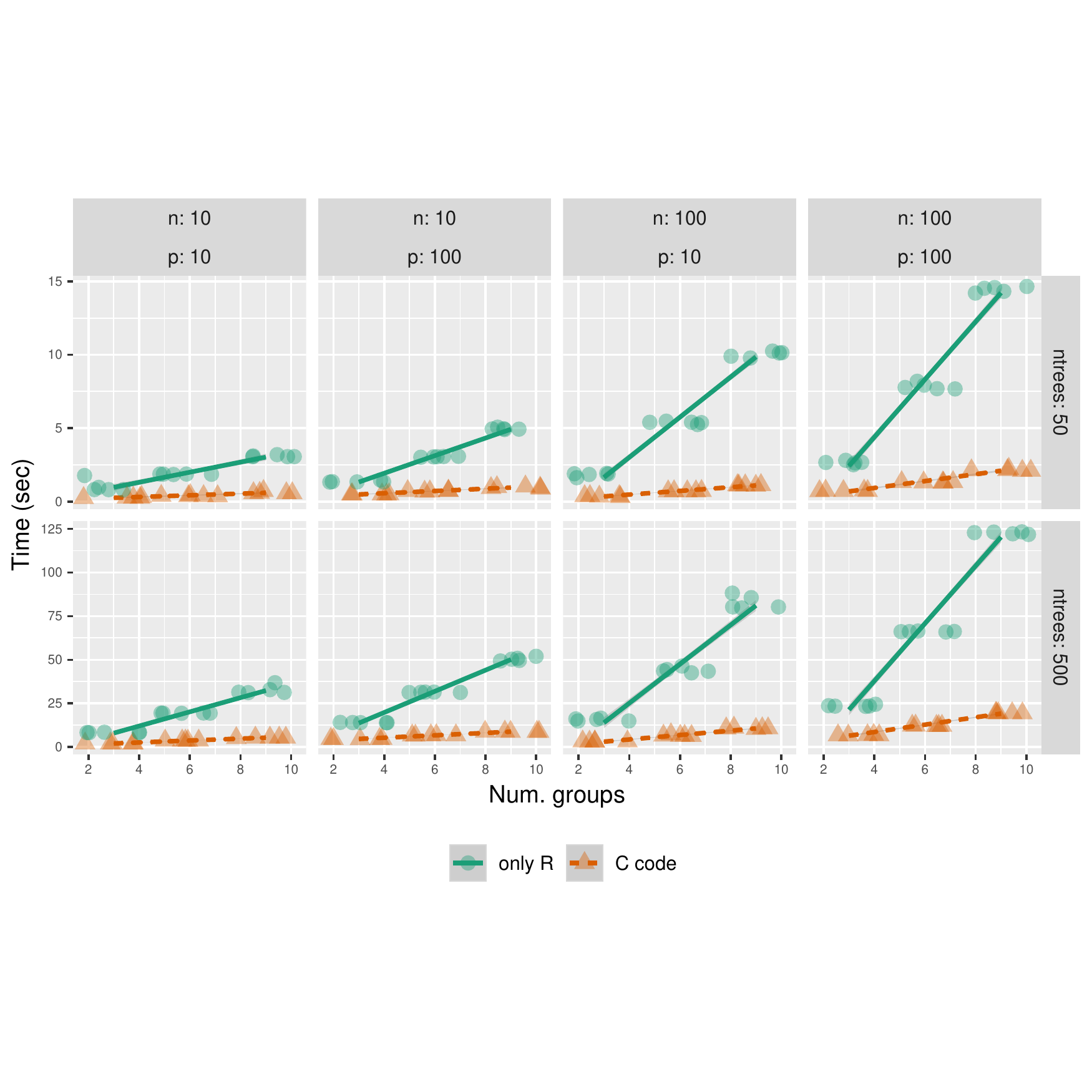} 

\end{knitrout}
\vspace{-3cm}
\caption{Computational performance for different sample sizes and number variables, with purely R code (green) and with C++ code (purple).\label{ratiotim}}
\end{figure}

The initial code for PPforest was developed entirely in R. It was subsequently profiled using \pkg{profvis}~\citep{profvis}, and two code optimization strategies were employed: translate main functions into \pkg{Rcpp} \citep{eddelbuettel2011rcpp} and parallelization  using \pkg{plyr}. The \pkg{microbenchmark} package was used to compare the speed before and after optimization. Figure \ref{ratiotim} shows the performance before and after optimization. The decrease in speed is linear as the number of groups increases. The improvement is between 3- and 9-fold for this range of parameters. The machine used for this comparison was a MacBook Pro with a processor of 2.4 GHz Intel Core i7 with a memory of 8GB and 1867MHz LPDDR3.

\begin{center}
\begin{table}[!ht]
\centering
\caption{Optimization assessment simulation design \label{respp1}}
\begin{tabular}{l|l}\hline\hline
  Parameters&Values\\ \hline\hline
 $g=$ number of classes &$(3, 3^2, 3^3)$\\
 $n=$ obs. by class & $(10^1, 10^2)$\\
 $p=$ number of variables & $(10^1, 10^2)$\\
 $m=$ number of trees & (50, 500)\\
 $cr=$ numbers of cores & $(1, 2, 4)$ \\
 $v$ PPforest version & (only~R, ~C~code)\\
\hline\hline
\end{tabular}
\end{table}
\end{center}

\subsection{PPF diagnostics}

The process of bagging and combining results from multiple trees produces numerous diagnostics which can provide a lot of insight into the class structure in high dimensions. Because ensemble methods are composed of many models fitted to subsets of the data, many statistics can be calculated to be analyzed as a separate data set. This provides the ability to understand how the model is working. The diagnostics of interest are the error rate, variable importance measure, vote matrix, and proximity matrix.

\subsubsection{Error rate}

Using the out-of-bag (oob) cases from bagged trees in the forest construction allows ongoing estimates of the generalization error for an ensemble of trees, described in \cite{breiman2001random}.
Given a training data set $d_n$, $B$ bootstrap samples from $d_n$ are taken. For each bootstrap sample ($b= 1, 2, \ldots B$), a \verb PPtree  classifier $h_n(\mathbf{x}, \Theta_b)$ is constructed, and a majority vote is used to get the PPF predictor.
The oob cases are used to get the error rate estimates. For each $\{\mathbf{x_i}, y_i\}$ in $d_n$, the votes are aggregated only for the classifiers $h_n(\mathbf{x}, \Theta_b)$ that do not contain $\{\mathbf{x_i}, y_i\}$. Hence, PPF is called the out-of-bag classifier, and the error rate for this classifier (out-of-bag error rate) is the estimate of the generalized error. The out-of-bag error rate is a measure for each model that is combined in the ensemble and is used to provide the overall error of the ensemble.

\subsubsection{Variable importance}

PPF calculates variable importance in two ways: (1) permuted importance using accuracy,  and (2) importance based on projection coefficients on standardized variables.
The permuted variable importance is comparable to the measure defined in the classical random forest algorithm. It is computed using the oob cases for the tree $k\;\;(B^{(k)})$ for each $X_j$ predictor variable.  Then the
permuted importance of the variable $X_j$ in the tree $k$ can be defined as:

\[
IMP^{(k)}(X_j) = \frac{\sum_{i \in B^{(k)} } I(y_i=\hat y_i^{(k)})-I(y_i=\hat y_{i,P_j}^{(k)})}{|B^{(k)}|}
\]

\noindent where $\hat y_i^{(k)}$
 is the predicted class for the observation $i$ in the tree $k$, and $y_{i,P_j}^{(k)}$ is the predicted class for the observation $i$ in the tree $k$ after permuting the values for variable $X_j$. The global permuted importance measure is the average importance over all the trees in the forest.
This measure is based on comparing the accuracy of classifying oob observations using the true class with permuted (nonsense) class.

For the second importance measure, the coefficients of each projection are examined. The magnitude of these values indicates importance if the variables have been standardized. The variable importance for a single tree is computed by a weighted sum of the absolute values of the coefficients across node, then the weights take the number of classes in each node into account($cl_{nd}$)~\citep{lee2013pptree} .
The importance of the variable $X_j$ in the PPtree $k$ can be defined as:

\[
IMP_{pptree}^{(k)}(X_j)=\sum_{nd = 1}^{nn}\frac{|\alpha_{nd}^{(k)}|}{cl_{nd} }
\]

\noindent where $\alpha_{nd}^{(k)}$ is the projected coefficient for node $ns$ and variable $k$ and $nn$ the total number of node partitions in the tree $k$.

The global variable importance in a PPforest then can be defined in different ways. The most intuitive are the average variable importance from each PPtree across all the trees in the forest.
\[
IMP_{ppforest1}(X_j)=\frac{\sum_{k=1}^K IMP_{pptree}^{(k)}(X_j)}{K}
\]
Alternatively, a global importance measure is defined for the forest as a weighted mean of the absolute value of the projection coefficients across all nodes in every tree. The weights are based on the projection pursuit indexes in each node ($Ix_{nd}$), and 1-(OOB-error of each tree)($acc_k$).

\[IMP_{ppforest2}(X_j)=\frac{\sum_{k=1}^K acc_k \sum_{nd = 1}^{nn}\frac{Ix_{nd}|\alpha_{nd}^{(k)}|}{nn }}{K}
\]

\subsubsection{Vote matrix}

An uncertainty measure for each observation, across models, is the proportion of times that a case is predicted to be in each class. If a case is always predicted to be the one class, there is no uncertainty about its group, and if this matches the true class then it is correctly labeled. Cases that are proportionately predicted to be multiple classes indicate difficult-to-classify observations. These cases may be important in that they might indicate special attention is needed in some neighborhoods of the data space, or more simply, could be errors in measurements in the data.

\subsubsection{Proximity matrix}

In a tree, each pair of observations can be in the same terminal node or not. Tallying this up across all trees in a forest gives the proximity matrix, an $n\times n$ matrix of the proportion of trees that the pair shares a terminal node. A proximity matrix can be considered to be a similarity matrix. This is typically used to do a follow-up cluster analysis to assess the strength of the class structure, and whether there are additional unlabeled clusters.

\subsubsection{Summary}

These diagnostics are used to assess model complexity; individual model contributions; variable importance and dimension reduction; and uncertainty in prediction associated with individual observations. 

\section{Performance comparison} \label{perfsec}

This section presents simulation results and a benchmark data study to examine the predictive performance of PPF in comparison to other methods. In the benchmark data study, PPF is compared with PPtree, CART and RF. The simulation results are designed to compare PPF with RF on data with linear projections defining class differences.

\subsection{Benchmark data study}

The performance of PPF is compared with the classification methods, PPtree, CART and RF using 10 benchmark data sets taken from the UCI Machine Learning archive \citep{Lichman}. Table \ref{bench.tab} presents summary information about the benchmark data, number of groups, cases, and predictors for each data set. The \textit{imbalance} between groups is measured by the range of group size proportions and \textit{correlation} is the average of all pairwise correlation coefficients among predictor variables.

\begin{table}[ht]
\centering
\caption{Summary of benchmark data. Imbalance and correlation indicating relative class sizes, and separations in combinations of variables.} 
\label{bench.tab}
\begin{tabular}{lrrrrr}
  \hline
DF & Cases & Predictors & Groups & Imbalance & Correlation \\ 
  \hline
crab & 200 &   5 &   4 & 0.00 & 0.95 \\ 
  lymphoma &  80 &  50 &   3 & 0.41 & 0.75 \\ 
  NCI60 &  61 &  30 &   8 & 0.07 & 0.56 \\ 
  parkinson & 195 &  22 &   2 & 0.51 & 0.50 \\ 
  fishcatch & 159 &   6 &   7 & 0.31 & 0.46 \\ 
  leukemia &  72 &  40 &   3 & 0.40 & 0.44 \\ 
  olive & 572 &   8 &   9 & 0.32 & 0.35 \\ 
  wine & 178 &  13 &   3 & 0.13 & 0.30 \\ 
  image & 2310 &  18 &   7 & 0.00 & 0.28 \\ 
  glass & 214 &   9 &   6 & 0.31 & 0.23 \\ 
   \hline
\end{tabular}
\end{table}

For each benchmark data set, $2/3$ of the observations are randomly chosen and used for training while the remaining $1/3$ are used as test data for computing predictive error. This procedure is repeated 200 times and the mean error rate is reported in Table \ref{respp1}. In PPF, the number of variables selected in each node partition is a tuning parameter, the proportion of variables selected at each partition. Three different values were used (0.6, 0.9 and the RF default). The test error reported for PPF is the best from these.

The results show that PPF has a better performance in the test data set than the other methods for the crab, fishcatch, leukemia, lymphoma, olive and wine data, while the RF test error is smaller for glass, image, NCI60 and parkinson data.

\begin{table}[ht]
\centering
\caption{Comparison of PPtree, CART, RF and PPF results with various data sets. The mean of training and test error rates from 200 re-samples is shown. (Order of rows is same as in Table 
ef{bench.tab}.) PPF performs favorably compared to the other methods. \label{res}} 
\begin{tabular}{l||cccc||cccc}
  \hline\hline &  \multicolumn{4}{c||}{TRAINING} & \multicolumn{4}{c}{TEST} \\ \hline
Data & CART & PPforest & PPtree & RF & CART & PPforest & PPtree & RF \\ 
  \hline
crab & 0.277 & 0.046 & 0.044 & 0.244 & 0.453 & 0.057 & 0.057 & 0.238 \\ 
  fishcatch & 0.118 & 0.000 & 0.000 & 0.193 & 0.184 & 0.011 & 0.012 & 0.191 \\ 
  glass & 0.237 & 0.306 & 0.331 & 0.240 & 0.330 & 0.390 & 0.403 & 0.224 \\ 
  image & 0.069 & 0.079 & 0.067 & 0.024 & 0.082 & 0.083 & 0.073 & 0.024 \\ 
  leukemia & 0.037 & 0.000 & 0.000 & 0.033 & 0.146 & 0.030 & 0.049 & 0.032 \\ 
  lymphoma & 0.052 & 0.000 & 0.000 & 0.100 & 0.155 & 0.053 & 0.069 & 0.081 \\ 
  NCI60 & 0.503 & 0.019 & 0.000 & 0.458 & 0.676 & 0.388 & 0.423 & 0.376 \\ 
  olive & 0.072 & 0.037 & 0.048 & 0.053 & 0.119 & 0.048 & 0.068 & 0.052 \\ 
  parkinson & 0.081 & 0.112 & 0.175 & 0.107 & 0.159 & 0.171 & 0.229 & 0.101 \\ 
  wine & 0.050 & 0.001 & 0.001 & 0.019 & 0.127 & 0.018 & 0.021 & 0.021 \\ 
   \hline
\end{tabular}
\end{table}

\begin{figure}[!hbpt]
\begin{knitrout}
\definecolor{shadecolor}{rgb}{0.969, 0.969, 0.969}\color{fgcolor}
\includegraphics[width=\maxwidth]{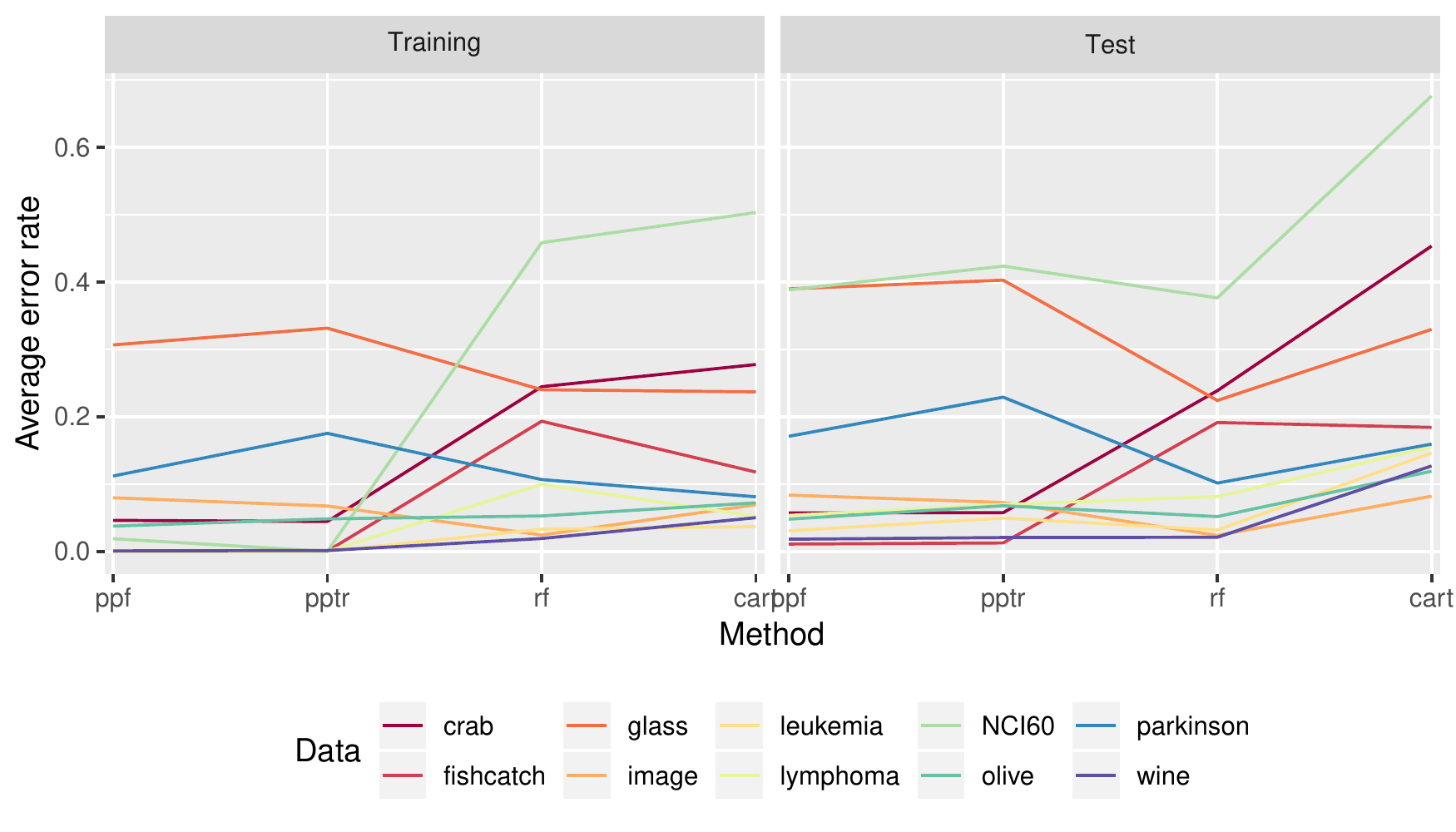} 

\end{knitrout}
\caption{Benchmark data results shown graphically. PPF performs consistently well across most of the data sets. \label{parallel}}
\end{figure}

Figures \ref{parallel} displays the performance comparison graphically. Each line connects the errors for one data set. Even though RF outperforms PPF on almost half the data (Table \ref{respp1}) PPF tends to have consistently low error.

\subsection{Boundary comparison with random forest}

To illustrate why and where PPF outperforms RF, results from a small simulation are shown. We expect PPF to outperform RF when the separation between classes is in linear combinations of variables. The simulated data is similar to the crab data.

\begin{figure}[!ht]
\begin{knitrout}
\definecolor{shadecolor}{rgb}{0.969, 0.969, 0.969}\color{fgcolor}
\includegraphics[width=\maxwidth]{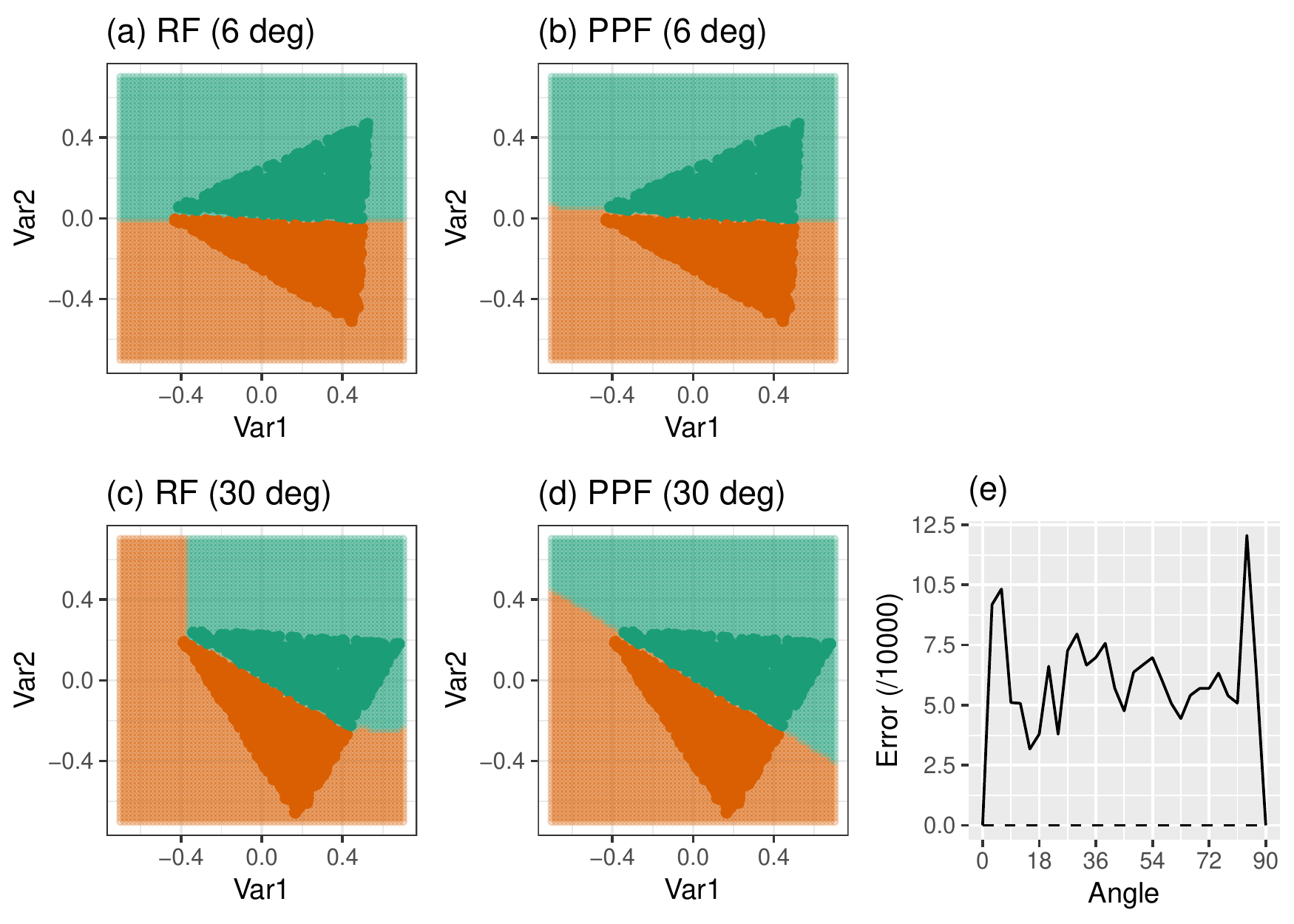} 

\end{knitrout}
\caption{Boundaries in rotated trangle simulation, for RF and PPF, for different rotations (a-d), and (e) the average error over 20 repetitions for different angle, solid line is RF, and dashed is PPF. PPF beats RF uniformly in this type of data, and RF produces inferior boundaries. A little surprising that RF does worse with a small rotation.}
\label{trianglesim}
\end{figure}

Each 2D simulated data set was rotated from 0 through 90$^o$, and 20 replications were conducted. Average (and standard deviation) of error was computed. Figure \ref{trianglesim} shows the boundaries for two of the rotations generated by the RF and PPF models, and shows the summary of the errors by rotation angle. PPF uniformly outperforms RF in this scenario and produces better boundaries.

\section{Diagnostics comparison}~\label{options}

The diagnostics computed by PPF (Section \ref{PPFsec}) and RF are compared for the lymphona data, which helps to understand why and how PPF outperforms RF with this data.

\subsection{Variable importance}

\begin{figure}[!ht]
\begin{center}
\begin{knitrout}
\definecolor{shadecolor}{rgb}{0.969, 0.969, 0.969}\color{fgcolor}
\includegraphics[width=\maxwidth]{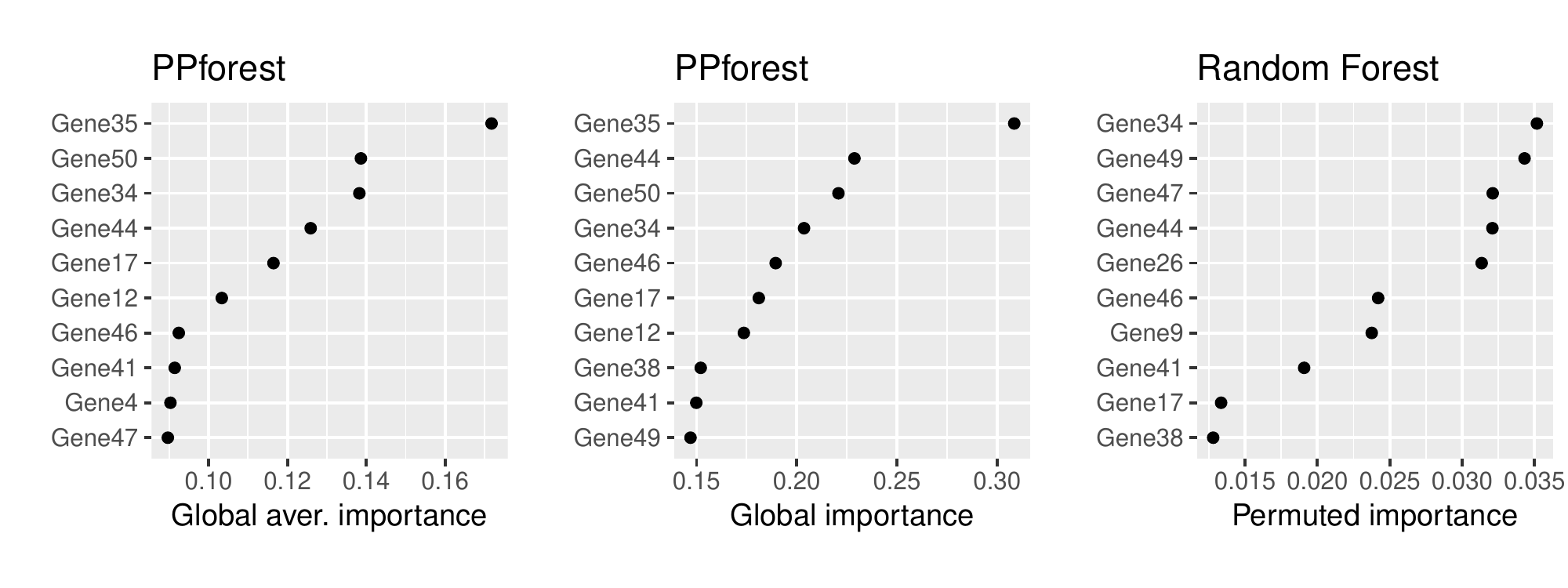} 

\end{knitrout}
\begin{knitrout}
\definecolor{shadecolor}{rgb}{0.969, 0.969, 0.969}\color{fgcolor}
\includegraphics[width=\maxwidth]{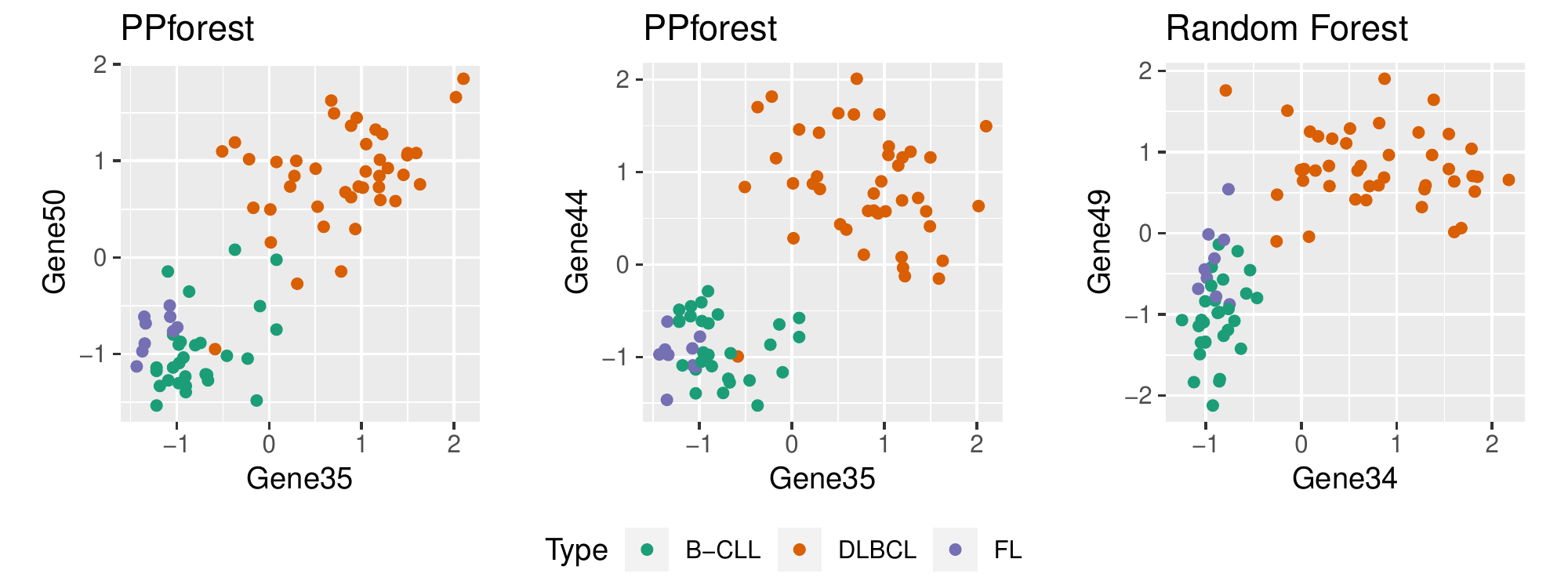} 

\end{knitrout}
\caption{Comparison of importance measures for the lymphoma data, where PPF outperformed RF. Top row shows the top 10 variables by each method, with two ways of calculating with PPF. Bottom row shows the top two variables from each, which illustrates the difference between methods. PPF is detecting differences between groups when there is association between variables. Using the global average importance (left), Gene35 and Gene50 better distinguish group FL. Using the global importance, Gene35 and Gene44 find a big gap between group DLBCL and the other two. \label{globalimp}}
\end{center}
\end{figure}

Figure \ref{globalimp} illustrates how the variable importance differs, using the lymphoma data. PPF outperformed RF for this data. There are three groups, and it is a high-dimension, low sample size data set. With PPF, the PDA index is used, and the 60\% of variables are available at each node. The number of trees used is the same as the RF default. Only the top ten most important variables are shown. There are some common on both lists and a some differences. Showing just the first two variables from each list is sufficient to illustrate the different type of boundaries induced by the classifiers. The two ways of computing importance in PPF do produce a different hierarchy of variables. With the global average importance, Gene35 and Gene50 are the top two, and these distinguish the small group FL best. With the global importance, Gene35 and Gene44 are featured, and together these find a big gap between DLBCL and the other two groups. PPF is utilizing the association between variables to classify groups, as would be expected.

\begin{figure}
\centering

\begin{knitrout}
\definecolor{shadecolor}{rgb}{0.969, 0.969, 0.969}\color{fgcolor}
\includegraphics[width=\maxwidth]{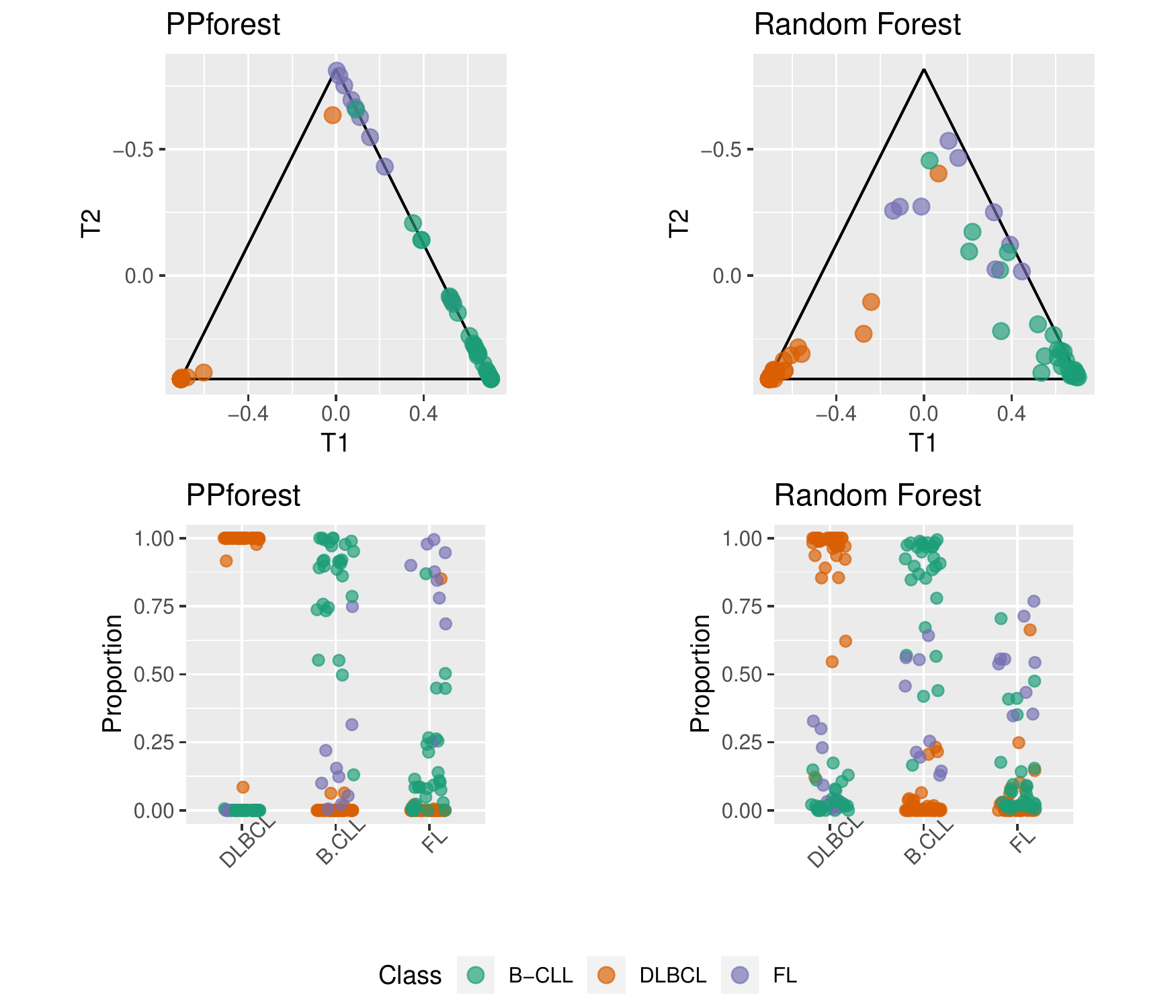} 

\end{knitrout}
\caption{Comparison of the out-of-bag vote  matrix for the three groups of the lymphoma data, returned by PPF and RF: (top) ternary plot, (bottom) side-by-side jittered dotplots. This illustrates the difference between methods. PPF votes more decidedly for most cases, than RF. Especially this is true for the DLCBL class, where all but one are almost always predicted to the true class.}
\label{voteplots}
\end{figure}

\subsection{Vote matrix}

Figure \ref{voteplots} shows the vote matrices returned by PPF and RF for three classes of the lymphona data. It is represented in two ways: as a ternary plot and as a side-by-side jittered  dotplot. The vote matrix has three columns corresponding to the proportion of times the case was predicted to be class B-CLL, DLBCL or FL, and thus is constrained to lie in a 2D triangle in 3D space. A ternary diagram is created using a helmert transformation of the vote matrix to capture the 2D subspace. The way to read it is: points near the vertex are clearly predicted to be one class, points along an edge are confused between two classes, and points in the middle are confused between the three classes. PPF provides more distinct classification of observations than RF, because the points are more concentrated in the vertices, and along one edge.

The side-by-side jittered dotplot is an alternative representation that readily can be used for any number of classes. The proportion each case is classified to a group is diplayed vertically along a horizontal axis representing the categorical class variable. Points are jittered a little horizontally to better see the distribution of proportions, and colour represents the true class. Points concentrated at the top part indicate cases that are clearly grouped into a class, and if the colour matches the true class then these are correct classifications. The message is similar to the ternary diagram: DLBCL is much more clearly distinguished by PPF, and FL is actually distinguishable from B-CLL by PPF but confused by RF.

\subsection{Proximity}

\begin{figure}[!h]
\centering
\begin{knitrout}
\definecolor{shadecolor}{rgb}{0.969, 0.969, 0.969}\color{fgcolor}
\includegraphics[width=\maxwidth]{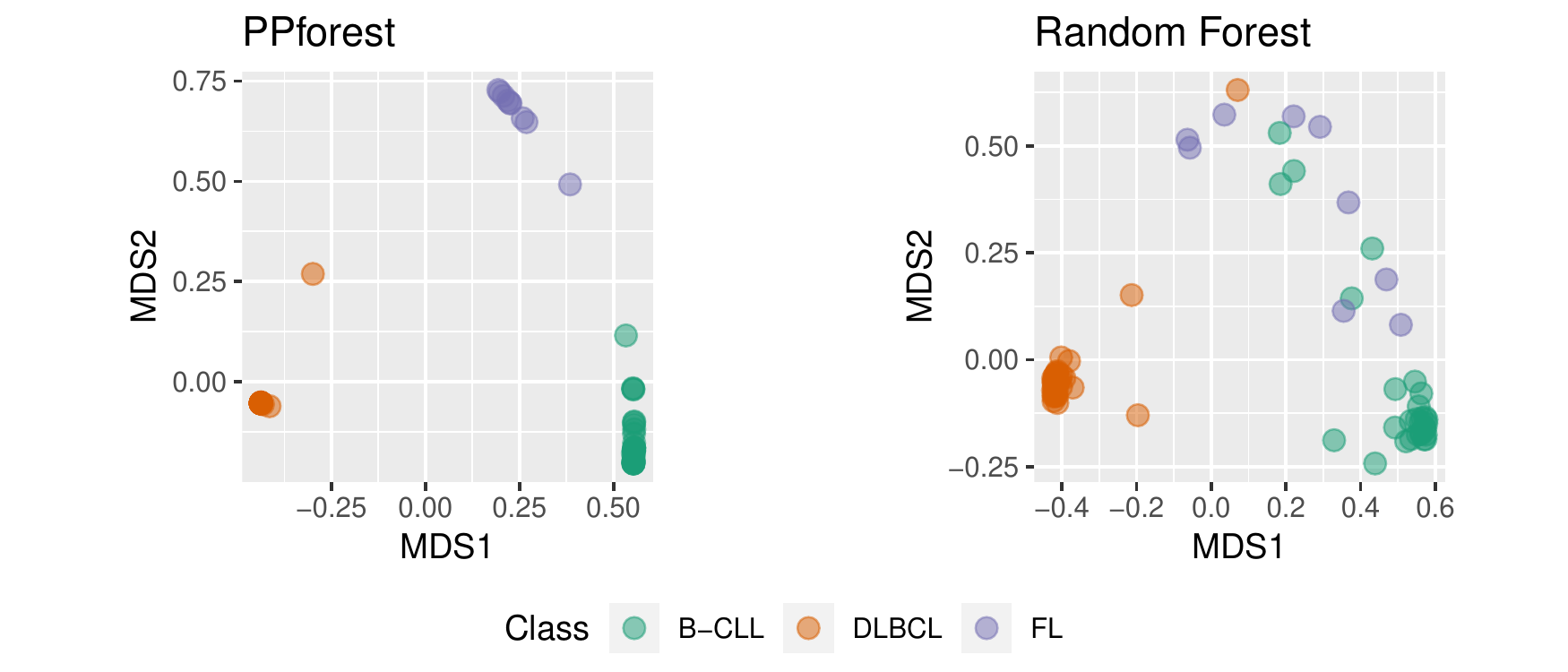} 

\end{knitrout}
\caption{Examining similarity between cases, using pairwise plots of multidimensional scaling on the proximity matrix from PPF and RF fits of the lymphoma data. It can be seen that most cases are grouped closely with their class in PPF while in RF FL and B-CLL are mixed. \label{prox1}}
\end{figure}

Figure \ref{prox1} shows multidimensional scaling plots of the proximity matrix produced by PPF and RF classification of the lymphoma data. PPF provides the cleaner proximities. This means that more frequently observations from the same class reside in the same terminal node of the trees making up the PPF, than those of RF.


\section{Parameter selection}

The primary parameters for PPF are mostly the same as those for RF: number of trees, and number of variables used in each node partition, with the addition of $\lambda$ when PDA is used as the index.

\begin{figure}[!h]
\centering
\begin{knitrout}
\definecolor{shadecolor}{rgb}{0.969, 0.969, 0.969}\color{fgcolor}
\includegraphics[width=\maxwidth]{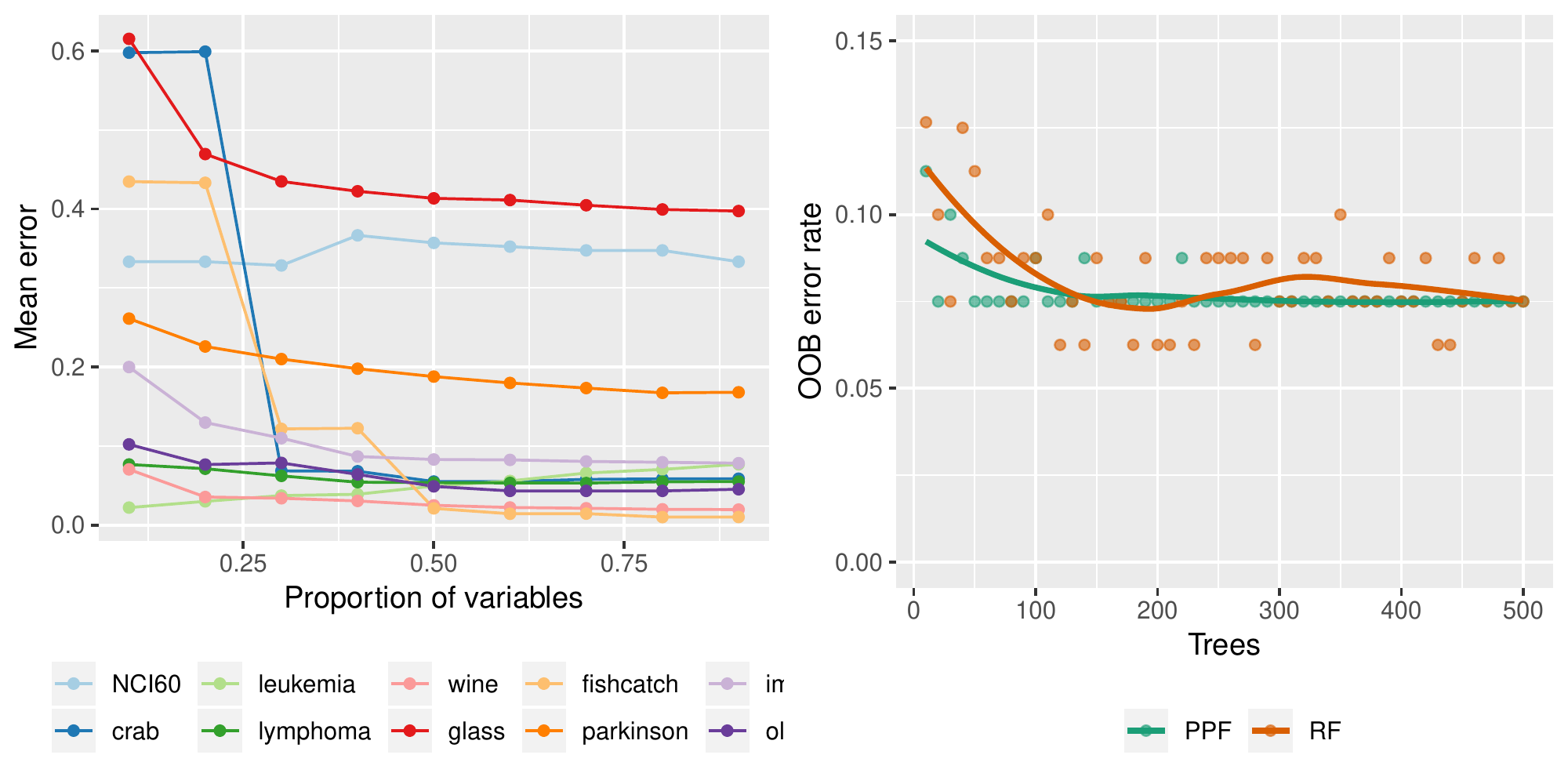} 

\end{knitrout}
\caption{Illustrating model tuning using error rate reduction. The average error rate plotted against proportion of variables in all the benchmark data is shown at left. The error rate tends to be better with more variables, but it does vary substantially by data set. OOB error is plotted against number of trees (right) on the lymphoma data for both PPF and RF. PPF has the consistently lower error, but both would indicate about 100 trees is sufficient to get the best results. \label{parameters} }
\end{figure}

Figure \ref{parameters} (left) shows the effect of proportion of variables for the benchmark data comparison. The average error over 200 training/test splits is shown. For all data sets error is lower when the more variables are used. Most converge to low error rate when half the variables are included.

The right plot compares the number of trees needed to optimise the OOB error for both PPF and RF on the lymphona data. Both need around 100 trees to produce best performance.

\begin{figure}[!ht]
\centering

\end{figure}

\section{Discussion}\label{discpp1}

This article has presented a new ensemble method (PPF) for classification problems, that is built on an oblique tree classifier (PPtree). PPF takes the correlation between variables into account. The forest algorithm enhances the single tree performance, adding diagnostics to assess variable importance, confusion of observations between groups and proximity of observations. It is best for medium sized data sets, both in number of observations and variables.

The benchmark data study showed that PPF predictive performance is always at least as good, or better, than CART and PPtree, and often better than RF. Simulation results show that PPF performs better than RF when the classes are separated by a linear combination of variables and when the correlation between variables increases. The variable importance diagnostic shows that different variables are combined to create the classification using a PPF than RF.

There are several directions where the work could be extended. The two projection pursuit indexes, LDA and PDA, can be readily supplemented by other indices. An example would be to add a regression index for a continuous response. Another direction is to adapt the PPtree algorithm to allow more than $g-1$ splits. This constraint protects the single tree model from overfitting. There is some protection against this with the bagging, and we expect it would enable deeper non-linear boundaries to be constructed by PPF. Lastly, because the accuracy of each tree is collected, automatic pruning of poor performing trees is a possibility.

\section{Acknowledgements}

The code for PPF are implemented in an R~\citep{RCore} package, \pkg{PPforest}, which is available on CRAN, with development versions at \url{https://github.com/natydasilva/PPforest}.

This paper was written with the R packages knitr~(\cite{xie:2015}), ggplot2 (\cite{hadley:2009}) and dplyr (\cite{dplyr}).

\bibliographystyle{spr-chicago}
\bibliography{ppfbiblio}

\end{document}